\documentclass[sigconf]{acmart}
\usepackage{multirow}
\usepackage{enumitem}   
\usepackage{algorithm}
\usepackage{algpseudocode}
\usepackage{tcolorbox}
\usepackage[table,xcdraw]{xcolor}
\usepackage{pifont}
\usepackage{bbding}
\newcommand{\ours}{TAROT}

\AtBeginDocument{%
  }

\copyrightyear{2026}
\acmYear{2026}
\setcopyright{cc}
\setcctype{by}
\acmConference[KDD 2026] {Proceedings of the 32nd ACM SIGKDD Conference on Knowledge Discovery and Data Mining V.2}{August 9--13, 2026}{Jeju Island, Republic of Korea.}
\acmBooktitle{Proceedings of the 32nd ACM SIGKDD Conference on Knowledge Discovery and Data Mining V.2 (KDD 2026), August 9--13, 2026, Jeju Island, Republic of Korea}
\acmISBN{979-8-4007-2259-2/2026/08}
\acmDOI{10.1145/3770855.3817944}

\begin{document}

\title{TAROT: Task-Adaptive Refinement of LLM-prior Graphs for Few-shot Tabular Learning}

\author{Ruxue Shi}
\authornote{Both authors contributed equally to this research.}
\orcid{0009-0007-9572-4018}
\affiliation{%
  \institution{Jilin University}
  \city{Changchun}
  \country{China}
}
\email{shirx24@mails.jlu.edu.cn}

\author{Yili Wang}
\authornotemark[1]
\orcid{0000-0003-0845-9521}
\affiliation{%
  \institution{Jilin University}
  \city{Changchun}
  \country{China}}
\email{wangyili@jlu.edu.cn}

\author{Mengnan Du}
\orcid{0000-0002-1614-6069}
\affiliation{%
  \institution{The Chinese University of Hong Kong, Shenzhen}
  \city{Shenzhen}
  \country{China}
}
\email{mengnandu@cuhk.edu.cn}

\author{Hangting Ye}
\orcid{0000-0001-6920-4181}
\affiliation{%
 \institution{Jilin University}
  \city{Changchun}
  \country{China}}
\email{yeht22@mails.jlu.edu.cn}

\author{Yi Chang}
\orcid{0000-0003-2697-8093}
\affiliation{%
  \institution{Jilin University}
  \city{Changchun}
  \country{China}}
\email{yichang@jlu.edu.cn}

\author{Xin Wang}
\orcid{0000-0001-9448-7689}
\authornote{Corresponding author.}
\affiliation{%
  \institution{Jilin University}
  \city{Changchun}
  \country{China}}
\email{xinwang@jlu.edu.cn}

\renewcommand{\shortauthors}{Ruxue Shi, Yili Wang, Mengnan Du, Hangting Ye, Yi Chang and Xin Wang}
\begin{abstract}
Few-shot tabular learning provides a cost-effective approach for real-world applications where annotation is costly and collecting sufficient samples for new tasks is difficult. Existing Traditional and LLM-based methods have demonstrated effectiveness in few-shot scenarios. However, traditional methods need additional training on unlabeled or generated data, which incur significant computational overhead. In addition, LLM-based methods that directly feed raw tabular data into LLMs raise privacy and compliance concerns. 
More importantly, both paradigms largely overlook the semantic relationships between features, which provide structural and semantic prior for constructing a \textbf{semantic graph}. 
Semantic graph is essential for modeling meaningful feature interactions in few-shot scenarios.
In this paper, we propose \ours, a GNN-based framework that encodes the structural and semantic prior by constructing and refining a task-adaptive semantic graph from this prior, thereby improving predictive performance in few-shot tabular learning. \ours\ first encodes heterogeneous tabular data into unified node semantic representations via a Unified Semantic Tabular Node Encoder (USTNE). 
Then, it prompts LLMs to infer the semantic relationship between features based on the task description and feature names to construct a semantic graph. 
To mitigate structural noise introduced by the hallucination of LLMs, \ours\ introduces Task-adaptive Semantic Graph Refinement that prunes spurious or task-unrelated edges and adds missing task-related ones, aligning the graph structure with the downstream objective. 
Finally, a GNN performs message passing over the refined graph to capture task-related semantic dependencies for prediction. Extensive experiments on various few-shot tabular learning benchmarks demonstrate the superior performance of \ours, establishing it as a state-of-the-art approach in this domain.


\end{abstract}





\maketitle
\newcommand\kddavailabilityurl{https://doi.org/10.5281/zenodo.20482615}
\ifdefempty{\kddavailabilityurl}{}{
\begingroup\small\noindent\raggedright\textbf{Resource Availability:}\\
The source code of this paper has been made publicly available at \url{\kddavailabilityurl}.
\endgroup
}
\begin{figure}[ht]
\centering
\includegraphics[width=0.49\textwidth]{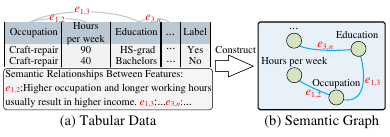}
\caption{\label{figure:motivation} Semantic graph construction on Adult dataset. (a) Semantic relationships between features on Tabular Data. (b) Semantic Graph for modeling meaningful feature interactions in few-shot scenarios.
} 

\end{figure}
\section{Introduction}
\begin{table*}[htbp]
  \centering
  \caption{Existing few-shot tabular learning methods can be broadly categorized into Traditional and LLM-based approaches. Most existing methods overlook the semantic relationships between features, and many of them are restricted to classification-only settings.}
  \resizebox{0.98\textwidth}{!}{
    \begin{tabular}{l|ccc|cccc|
    c}
    \toprule
    \multicolumn{1}{c|}{\multirow{2}[4]{*}{Properties}} & \multicolumn{3}{c|}{Traditional Few-shot Tabular Learning Methods} & \multicolumn{4}{c|}{LLM-based Few-shot Tabular Learning Methods} & \multirow{2}[4]{*}{\textbf{\ours\ (Ours)}} \\
\cmidrule{2-8}          & SCARF~\cite{bahri2021scarf}  & TabPFN~\cite{hollmann2022tabpfn} & STUNT~\cite{nam2023stunt} & In-context~\cite{wei2022emergent} & TABLET~\cite{slack2023tablet} & TabLLM~\cite{hegselmann2023tabllm} & FeatLLM~\cite{featllm} &  \\
    \midrule
       No additional training required & \XSolidBrush(Contrastive Learning)   &  \XSolidBrush(Supervised Learning)    & \XSolidBrush(Meta learning)      & \cellcolor{red!20}\CheckmarkBold     & \cellcolor{red!20}\CheckmarkBold     & 
       \XSolidBrush(Fine-tune LLM) &\cellcolor{red!20}\CheckmarkBold      & \cellcolor{red!20}\CheckmarkBold \\
    No LLM access sample required &\cellcolor{red!20}\CheckmarkBold      & \cellcolor{red!20}\CheckmarkBold     &\cellcolor{red!20}\CheckmarkBold      &  \XSolidBrush     & \XSolidBrush      & \XSolidBrush      & \XSolidBrush      &\cellcolor{red!20}\CheckmarkBold  \\
    Classification task &\cellcolor{red!20}\CheckmarkBold      & \cellcolor{red!20}\CheckmarkBold     &\cellcolor{red!20}\CheckmarkBold      &\cellcolor{red!20}\CheckmarkBold      &\cellcolor{red!20}\CheckmarkBold     &\cellcolor{red!20}\CheckmarkBold      &\cellcolor{red!20}\CheckmarkBold      & \cellcolor{red!20}\CheckmarkBold \\
    Regression task & \cellcolor{red!20}\CheckmarkBold     &  \XSolidBrush     &   \XSolidBrush    &   \cellcolor{red!20}\CheckmarkBold   &  \XSolidBrush    & \XSolidBrush      &   \XSolidBrush    & \cellcolor{red!20}\CheckmarkBold \\
    Semantic relationships between features & \XSolidBrush       & \XSolidBrush      & \XSolidBrush      &\XSolidBrush        & \XSolidBrush       & \XSolidBrush      &\XSolidBrush       & \cellcolor{red!20}\CheckmarkBold \\
    \bottomrule
    \end{tabular}%
    }
  \label{tab:few_shot}%
\end{table*}%
Given the substantial financial and temporal costs of sample annotation~\cite{clements2020sequential,nam2023stunt} and the difficulty of collecting data for new tasks (e.g., some rare or new diseases)~\cite{mitani2020small,mondal2020data}, learning from a limited number of labeled samples has emerged as a cost-effective solution for real-world deployment of machine learning models~\cite{snell2017prototypical,wang2020generalizing,oreshkin2018tadam,wang2021structure}. This scenario, commonly referred to as \textbf{few-shot learning}, has recently attracted increasing attention across multiple domains, including computer vision~\cite{chen2019closer,peng2019few} and tabular learning~\cite{nam2023stunt,featllm}. However, with insufficient supervisory signals, traditional supervised learning struggles to learn effective models, as its performance heavily relies on statistical convergence over large labeled datasets. This limitation is particularly pronounced in tabular learning, where labeled data are often scarce~\cite{liud2r2} in real-world applications, such as fraud detection~\cite{frauddetection} and disease diagnosis~\cite{diseasediag}.

To tackle such limited label issues, existing few-shot learning approaches for tabular data can be broadly classified into two categories. \textit{Traditional methods} aim to acquire transferable representations or useful knowledge by additional training on large-scale unlabeled or synthetic tabular data. For instance, SCARF~\cite{bahri2021scarf} and STUNT~\cite{nam2023stunt} leverage unlabeled tabular data to respectively learn a generalizable and adaptable representation, while TabPFN~\cite{hollmann2022tabpfn} is trained on large-scale generated datasets to incorporate prior knowledge over feature–label relationships, enabling rapid adaptation in few-shot scenarios. In contrast, \textit{LLM-based methods} transform raw tabular samples into natural language representations and exploit the inherent knowledge of LLMs~\cite{yao2024survey,bubeck2023sparks} for in-context reasoning~\cite{wei2022emergent,slack2023tablet}, feature importance estimation~\cite{featllm}. Furthermore, it uses task-specific fine-tuning to improve LLM capabilities for tabular understanding and downstream performance~\cite{hegselmann2023tabllm}.

Despite recent advances in these approaches, they still suffer from some limitations that hinder their effectiveness and scalability in real-world deployment, as summarized in Tab.~\ref{tab:few_shot}. 
Traditional methods incur substantial computational overhead when trained on large-scale unlabeled or synthetic tabular data~\cite{hollmann2022tabpfn}. On the other hand, LLM-based methods are constrained by the context length of LLMs~\cite{wang2024beyond,an2024does}, and sending raw data to external models raises the concern about privacy and compliance~\cite{carlini2021extracting,kim2023propile}. More importantly, both paradigms largely overlook the semantic relationships between features as shown in Fig.~\ref{figure:motivation} (a), which provide structural and semantic prior for constructing a \textbf{semantic graph}, like Fig.~\ref{figure:motivation} (b). This semantic graph enables the modeling of meaningful feature interactions, addressing the instability and susceptibility to spurious correlations that arise when feature interactions are learned directly from sparse supervision in few-shot scenarios.

However, to the best of our knowledge, no prior work has successfully integrated semantic graphs into few-shot tabular data learning, primarily due to two key challenges: \ding{182} \textbf{Difficulty in obtaining graph structures.} The semantic relationships between features of tabular data are often not explicitly provided, which makes the corresponding semantic graph structure unavailable.~\cite{guo2021tabgnn}. Meanwhile, existing graph structure learning methods~\cite{liao2023tabgsl,yan2023t2g} typically require a large amount of data to accurately infer semantic relationships between features, which limits their effectiveness in few-shot scenarios. \ding{183} \textbf{Structure noise in the semantic graph.} This noise arises from spurious or task-unrelated edges and missing task-related ones. In few-shot scenarios, such noise can misguide message passing and amplify irrelevant correlations, leading to unreliable predictions~\cite{dai2022towards}.

In this paper, we propose \ours, a \underline{\textbf{T}}ask-\underline{\textbf{A}}daptive \underline{\textbf{R}}efinement \underline{\textbf{o}}f LLM-prior Graphs for Few-shot \underline{\textbf{T}}abular Learning. \ours\ is a GNN-based framework that improves prediction by explicitly modeling feature interactions through task-adaptive semantic graph construction and refinement. Our key innovation lies in constructing and refining a task-adaptive semantic graph from the structural and semantic prior under limited data. The semantic graph (i) emphasizes meaningful feature interactions. and (ii) mitigates the adverse effects of structural noise. This enables more reliable message passing and improves predictive performance. Specifically, \ours\ first introduces a Unified Semantic Tabular Node Encoder (USTNE) that encodes heterogeneous tabular features into unified node semantic representations using a pre-trained encoder. Next, it prompts LLMs to infer semantic relations between features based on the task objective description and feature names to an initial semantic graph (Challenge \ding{182}). Then, we refine this graph in a task-adaptive manner by pruning spurious and task-unrelated edges and adding missing task-related semantic ones, thereby denoising the LLM-induced structural noise (Challenge \ding{183}) and generating a task-adaptive semantic graph. Finally, we apply a GNN over the refined graph to model feature interactions, thereby capturing semantic dependencies that benefit downstream prediction. Our main \textbf{contributions} are summarized as follows: 

\begin{itemize}[noitemsep,leftmargin=*,label=$\star$]
   \item We propose a novel insight that leverages LLMs to provide structural prior knowledge by inducing a semantic graph based on the task description and feature names, addressing the challenge that requires a large amount of data to accurately infer semantic relationships between features.
    \item We introduce a task-adaptive refinement mechanism that denoises semantic graphs by removing task-unrelated edges and adding missing task-related ones, enabling effective GNN-based feature interaction modeling for few-shot tabular learning.
   \item Extensive experiments on 11 real-world datasets show that \ours\ consistently outperforms state-of-the-art baselines, while quantitative and qualitative analyses confirm the effectiveness of the generated task-adaptive semantic graphs for few-shot tabular learning.
\end{itemize}

\begin{figure*}[t]
\centering
\includegraphics[width=1.0\textwidth]{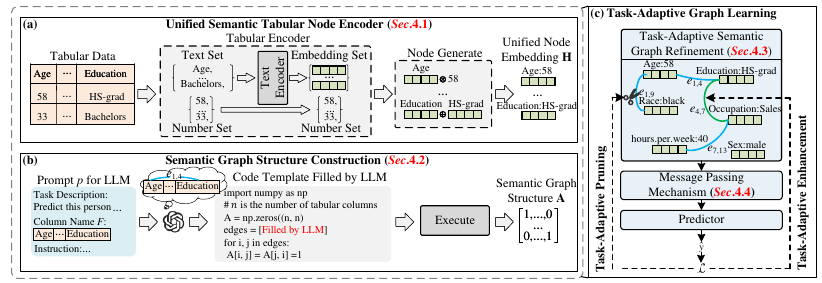}
\caption{\label{figure:framework} Overview of \ours. USTNE encodes heterogeneous tabular data into unified node representations. Then, LLMs construct a semantic graph based on task descriptions and feature names, which is refined via Task-Adaptive Semantic Graph Refinement to reduce noise introduced by LLMs. A GNN finally encodes the refined graph for prediction.
}
\end{figure*}

\section{Related Work}
\subsection{Few-shot Tabular Learning}
Few-shot tabular learning has been proposed to address scenarios where annotations are costly, and data for emerging tasks (e.g., some rare or new diseases) are scarce~\cite{nam2023stunt}. Recent advances in this area can be broadly categorized into two categories: \textit{Traditional few-shot tabular learning methods} and \textit{LLM-based few-shot tabular learning methods}. Traditional approaches, such as SCARF~\cite{bahri2021scarf}, STUNT~\cite{nam2023stunt}, and TabPFN~\cite{hollmann2022tabpfn}, leverage large-scale unlabeled data or synthetic datasets for additional training, thereby capturing transferable tabular patterns that improve downstream performance under limited supervision. In parallel, LLMs, trained on massive real-world corpora, encode substantial world knowledge~\cite{hu2025enabling,hou2024wikicontradict} and exhibit strong reasoning capabilities~\cite{laban2023summedits,wan2023better}, making them increasingly attractive for tabular learning. Existing LLM-based approaches typically serialize tabular data into natural language representations and prompt LLMs to perform tabular prediction tasks. For example, TABLET~\cite{slack2023tablet} enhances tabular reasoning by incorporating task-specific instructions into prompts, while FeatLLM~\cite{featllm} employs LLMs as feature engineers to automatically filter important features before training downstream classifiers. Alternatively, TabLLM~\cite{hegselmann2023tabllm} fine-tunes LLMs on tabular data to improve their capability in understanding and processing tabular data. Despite their effectiveness, most existing methods overlook semantic graph structures, whose message passing mechanisms capture semantic dependencies and improve few-shot predictions.
\subsection{Graph-based Tabular Learning}
Tabular data often exhibit a semantic graph structure, i.e., semantic relationships among features. Such semantic structures model meaningful feature interactions and improve predictive performance.~\cite{yan2023t2g}. One straightforward way to obtain a tabular graph is manual feature engineering~\cite{seide2011feature}. For example, TabGNN~\cite{guo2021tabgnn} constructs a multigraph using multiple hand-designed relational rules and performs message passing to enhance tabular prediction. However, manual graph construction typically requires domain expertise and incurs high human cost. To reduce this burden, recent work explores semantic graph structure learning, enabling models to autonomously infer semantic relations among features. For instance, TabGSL~\cite{liao2023tabgsl} explicitly initializes the learner-view graph with kNN and preserves top-$k$ connections for graph construction. In addition, T2G-FORMER~\cite{yan2023t2g} derives feature relationships by computing similarity scores, converting them into graph structures, and modeling them with graph transformers to capture semantic dependencies. While these approaches can learn effective semantic graphs given sufficient training data, they often struggle in few-shot scenarios where supervision is limited, making reliable semantic graph structure construction challenging. To address this, we propose \ours, which prompts LLMs to infer semantic relations among features based on the task objective and feature names, forming an initial semantic graph. We then refine this graph in a task-adaptive manner by pruning spurious edges and adding missing task-relevant ones, resulting in a meaningful task-related semantic graph.


\vspace{-14pt}
\section{Preliminaries}
\subsection{Problem Definition}
Given a tabular dataset $D=\{\mathbb{X}, \mathbb{Y}\}$, each tabular sample $x \in \mathbb{X}$ is represented as tabular feature $\{x_i\}_{i=1}^n$, where \(n\) is the number of tabular features that compose of $N_n$ numerical features and $N_c$ categorical features. The corresponding column name of each feature is $f_i \in F$. The goal is to train a model \(g_\theta : \mathbb{X} \to \mathbb{Y} \), parameterized by \( \theta \), to map the row feature space \( \mathbb{X} \) to the label space \( \mathbb{Y} \), where for binary classification \( \mathbb{Y} =\{0,1\}\), for multiclass classification  \( \mathbb{Y} =\{1,\ldots,c\}\), and for regression \( \mathbb{Y} =\mathbb{R}\).
We study the \(k\)-shot tabular learning by restricting supervision to \(k\) labeled samples during training.

\vspace{-10pt}
\subsection{Semantic Graph of Tabular Data}
\begin{definition}[Semantic Graph of Tabular Data]
Tabular data exhibits an implicit semantic graph \(\mathcal{G}=(\mathcal{V},\mathcal{E},\textbf{A},\textbf{H})\), where each node $v_i\in\mathcal{V}$ corresponds to a \textbf{tabular feature \(x_i\)}, and an undirected edge \(e_{ij}=\{v_i,v_j\}\in \mathcal{E}\) indicates that there exist \textbf{semantic relationships} between tabular features \(x_i\) and \(x_j\), where the semantics are primarily provided by the corresponding \textbf{column names} \(f_i\) and \(f_j\). \(\textbf{A}\in\{0,1\}^{n\times n}\) is the adjacency matrix, where \(\textbf{A}_{ij}=1\) if \(\{v_i,v_j\}\in \mathcal{E}\), and \(\textbf{A}_{ij}=0\) otherwise, and \(\textbf{H} \in \mathbb{R}^{n\times d}\) contains a \(d\)-dimensional embedding for each node. \end{definition}

The semantic graph of tabular data facilitates the modeling of meaningful feature interactions, alleviating the instability and susceptibility to spurious correlations that arise when feature interactions are learned directly from sparse supervision in few-shot scenarios. However, in the real-world tabular dataset, semantic relationships between features of tabular data are often not explicitly provided. It makes the semantic graph structure \(\mathcal{E}\) and corresponding adjacency matrix \(\textbf{A}\) unavailable.

\section{Proposed Method: \ours}
In this section, we introduce the overall framework of \ours, as illustrated in Fig.~\ref{figure:framework}.
\ours\ constructs a task-adaptive semantic graph to facilitate task-aware message passing across related features, thereby stabilizing representation learning and improving predictive performance in few-shot tabular scenarios.
The framework consists of four main components: (i) Unified Semantic Tabular Node Encoder (USTNE) (Sec.~\ref{sub:hne}). (ii) Semantic Graph Structure Construction, 
(Sec.~\ref{sub:cod}). (iii) Task-Adaptive Semantic Graph Refinement (Sec.~\ref{sec:task-adaptive}). (iv) Message Passing Mechanism (Sec.~\ref{sec:mpm}).

\subsection{Unified Semantic Tabular Node Encoder}
\label{sub:hne}
Tabular data typically comprises heterogeneous features, which poses a major challenge for existing embedding methods, especially in few-shot scenarios, where models are more prone to overfitting. To tackle this issue, we propose USTNE, a training-free unified semantic tabular data encoder that transforms the heterogeneous tabular sample $x=\{x_i\}_{i=1}^n$ into a unified semantic node representation matrix \(\textbf{H}=\{h_i\}_{i=1}^n\). We first decompose the tabular data into a text set \(\text{\textbf{Text}}\) and a numerical set \(\text{\textbf{Num}}\) as follows:
\begin{equation}
\text{\textbf{Text}}, \ \text{\textbf{Num}} = \mathrm{Extract}\left(D, F\right),
\end{equation}
where, the function \(\mathrm{Extract}(\cdot)\) collects all elements from the tabular data \(D\) and feature names \(F\), and then assigns them to the textual subset \(\text{\textbf{Text}}\) and the numerical subset \(\text{\textbf{Num}}\), based on their data types. For example, as shown in Fig.~\ref{figure:framework} (a), categorical types such as Age, Education, and HS-grad are assigned to \textbf{Text}, whereas numerical types (e.g., 58 and 33) are assigned to \textbf{Num}. To ensure stable optimization, numerical features are normalized to mitigate scale discrepancies before being fed into the model. Finally, since existing pre-trained encoders (e.g., BERT~\cite{devlin2019bert}) have relatively weak encoding capabilities for numerical data, we only use a pre-trained text encoder to encode the textual subset \(\text{\textbf{Text}}\):
\begin{equation}
\text{\textbf{T}} = \mathrm{Text\_Encoder}(\text{\textbf{Text}}),
\end{equation}
where \(\text{\textbf{T}} \in \mathbb{R}^{n_t\times d}\) is the set of text embedding, \(n_t\) is the number of text element in tabular data \(D\). Finally, for categorical features, both representation of the feature name and feature value are retrieved from \(\text{\textbf{T}}\). For numerical features, representation of the feature name is retrieved from \(\text{\textbf{T}}\), while the numerical value is retrieved from \(\text{\textbf{Num}}\). The following procedure is then applied to obtain node representations that incorporate semantic information in the form of key value pair (\(f_j\),\(x_{i}\)):
\begin{equation}
h_{i} =
\begin{cases} 
\text{\textbf{T}}_{f_i} \oplus \text{\textbf{T}}_{x_{i}}, & \text{for categorical feature}, \\[3pt]
\text{\textbf{T}}_{f_i} \odot \text{\textbf{Num}}_{x_{i}}, & \text{for numerical feature},
\end{cases}
\end{equation}
where \(h_{i} \in \mathbb{R}^{d}\) denotes the node embedding of the \(i\)-th feature, \(\oplus\) represents the concatenation operation, and \(\odot\) represents element-wise multiplication between the embedding of the feature name and the corresponding value. Through the aforementioned encoding process, we have transformed the heterogeneous tabular sample $x=\{x_i\}_{i=1}^n$ into a unified semantic nodes representation \(\textbf{H}=\{h_i\}_{i=1}^n\) in the semantic graph. 

\subsection{Semantic Graph Structure Construction}
\label{sub:cod}

\begin{figure}[t]
\centering
\includegraphics[width=0.45\textwidth]{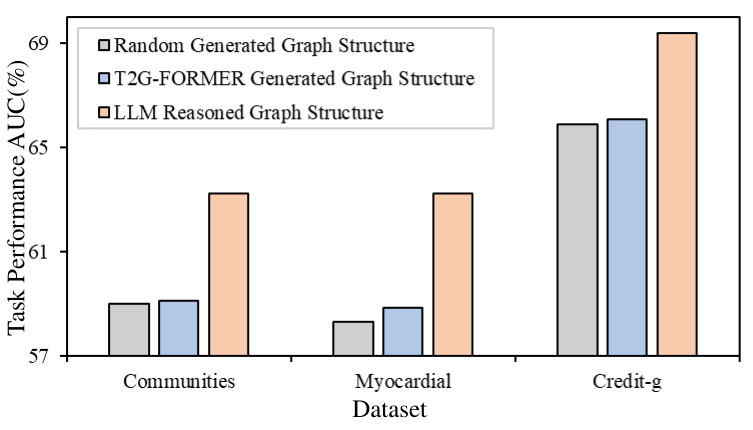}
\caption{\label{figure:method} Comparison of the impact of graph structures generated by different methods on task performance at 16-shot setting. For a fair comparison, the three methods differ only in their graph structures.
}
\end{figure}
To address Challenge \ding{182} (i.e., the difficulty of constructing semantic graphs for tabular data in few-shot scenarios), we leverage LLMs as an external knowledge source. Benefiting from real-world knowledge acquired during pretraining and strong zero-shot reasoning ability, LLMs can capture semantic relationships between features and thus facilitate semantic graph construction. Our objective is to use LLMs to generate a semantic graph structure $\mathcal{E}$ based on the task description and feature names. Specifically, we construct a prompt \(p\) for the LLM to guide it to understand the structure of tabular data and infer semantic relationships between features by leveraging its diverse, aggregated knowledge. To better exploit the LLM’s reasoning ability and to standardize its output, the prompt \(p\) (please see Appendix~\ref{suba:alg} Tab.~\ref{tab:adult}) consists of three components: (i) \(p_{meta}\) (ii) \(p_I\) (iii) \(p_{code}\), where, Meta Information \(p_{meta}\) describes the task objective and the feature names; Instruction \(p_I\) constrains the model to infer semantic relationships between features in a step-by-step (i.e. CoT) manner, while standardizing LLM by filling in the parts related to semantic relationship between features in the \(p_{code}\). Instead of first getting the edge set $\mathcal{E}$ and then converting it into the adjacency matrix \textbf{A}, we prompt the LLM to generate an \emph{executable} code snippet that directly outputs the adjacency matrix \textbf{A} by filling the code template in Fig.~\ref{figure:framework}(b).  This design avoids additional parsing and post-processing, leading to a more deterministic and implementation-friendly graph structure construction:
\begin{equation}
\begin{split}
    \textbf{A}&= \text{Execute}(\text{LLM}(p)) \\&= \text{Execute}(\text{LLM}(p_{meta} \oplus p_I \oplus p_{code})),
\end{split}
\end{equation}
where \(\oplus\) is the concatenation of each prompt, and this process is achieved through querying the LLM, without requiring any fine-tuning. To validate the effectiveness of the LLM-generated semantic graph structure, we compare it against (i) randomly generated graphs and (ii) the neural baseline T2G-FORMER~\cite{yan2023t2g}. As shown in Fig.~\ref{figure:method}, the LLM-generated graph yields significant improvements on downstream tasks under the 16-shot setting. Notably, the LLM requires only basic meta information (e.g., feature names and task description) and does not access any data samples. Therefore, graph construction is performed in a zero-shot manner, which mitigates potential privacy risks associated with sending raw data to the LLM. Despite the LLM-generated semantic graph structures yielding significant gains on downstream tasks, we observe an issue (Challenge \ding{183}) illustrated in Fig.~\ref{figure:framework} (c): Because of hallucination problems~\cite{hallucination,hallucination2,li2024mitigating}, LLMs may introduce spurious edges that are irrelevant to the downstream task and may also miss task-related semantic ones. For example, for a task that predicts whether an individual’s income exceeds \$50{,}000, LLMs might incorrectly connect age and race, edges that are not useful for the task, while omitting necessary connections among truly semantic informative features. In few-shot scenarios, such structure noise can misguide message passing and amplify irrelevant correlations, leading to unreliable predictions. To address this, we treat the LLM-generated semantic graph structure \(\textbf{A}\) as a prior graph structure \(\textbf{A}_{prior}\) and further refine it via Task-Adaptive Semantic Graph Refinement.

\subsection{Task-Adaptive Semantic Graph Refinement}
\label{sec:task-adaptive}
To address Challenge \ding{183} (i.e., structural noise introduced by LLM), we propose a task-adaptive framework that refines the graph structure \(\textbf{A}_{prior}\) using only a small set of labeled samples via task-adaptive graph optimization. Specifically, we prune spurious and task-unrelated edges and add missing task-related semantic ones to denoise the LLM-induced structural noise and obtain a structure that is consistent with the downstream task. In the few-shot scenarios, learning a graph solely from limited supervision is often unstable and yields suboptimal structures. By anchoring optimization around \(\textbf{A}_{prior}\), our method avoids an expensive search over the entire structure space and instead performs localized, semantics-preserving adjustments to achieve task-adaptive refinement. We first construct edge representation for each node pair (\(v,u\)) by combining their node representations, as follows:
\begin{equation}
    e_{v,u}=\big[(h_v,h_u) \oplus (h_v\odot h_u)\oplus(|h_v-h_u|)\big],
\end{equation}
where \(\oplus\) represents the concatenation operation, and \(\odot\) represents element-wise multiplication. We construct reliable semantic edge representation \(e_{v,u} \in \mathbb{R}^{3d}\) by concatenating \((h_v, h_u)\), their element-wise product \((h_v \odot h_u)\), and absolute difference \((|h_v-h_u|)\). Next, we use a linear model to score the edge representations \(e_{v,u}\) as follows:
\begin{equation}
    s_{v,u} = \sigma(e_{v,u}w),
\end{equation}
where \(s_{v,u}\) is the semantic score of the edge \((v, u)\), \(\sigma\) is the Sigmoid function, and \(w \in \mathbb{R}^{3d\times1}\) represents the trainable weight in the linear model. To learn structural information from the prior semantic graph \(\textbf{A}_{prior}\), we impose a prior learning loss, as follows:
\begin{equation}
\label{eq:7}
\mathcal{L}_{prior}
= \frac{1}{|\Omega|}\sum_{(v,u)\in \Omega}
\log\!\left(1+\exp(-s_{v,u})\right),
\end{equation}
\begin{equation}
    \Omega=\{(v,u)\mid A_{\text{prior}}(v,u)=1\}.
\end{equation}

Overall, this loss encourages the model to regularize toward the prior graph structure \(\textbf{A}_{prior}\) and inject the structural knowledge encoded in \(\textbf{A}_{prior}\) into the learning process, thereby mitigating overfitting caused by scarce labeled data in few-shot scenarios. Then, based on the semantic scores, we derive the set of task-unrelated edges to prune and the set of task-related ones to add as follows:
\begin{equation}
\label{eq:prune}
\textbf{A}_{prune}(v,u) =
\begin{cases} 
0, & s_{v,u}\geq\tau, \\[3pt]
1, & s_{v,u}<\tau,
\end{cases}
\end{equation}

\begin{equation}
\label{eq:enhance}
\textbf{A}_{enhance}(v,u) =
\begin{cases} 
0, & s_{v,u} \notin \text{Topk}(\{ s_{v,u}\}_{\substack{1\le v,u \le n}}), \\[3pt]
1, & s_{v,u} \in \text{Topk}(\{s_{v,u}\}_{\substack{1\le v,u \le n}}),
\end{cases}
\end{equation}
where \(\textbf{A}_{prune} \in \mathbb{R}^{n \times n}\) denotes a binary mask of task-unrelated edges to be removed, obtained by marking edges whose semantic scores fall below a threshold \(\tau\). \(\textbf{A}_{enhance} \in \mathbb{R}^{n \times n}\) denotes a binary mask of task-related edges to be added, obtained by selecting the topk edges with the highest semantic scores. Based on these two masks, we refine the prior graph \(\textbf{A}_{prior}\) by first pruning noisy connections and then adding missing relations:
\begin{equation}
    \textbf{A}_{refine} = (\textbf{A}_{prior} \odot (1-\textbf{A}_{prune})) \lor \textbf{A}_{enhance},
\end{equation}
where, \(\textbf{A}_{refine}\) represents task-adaptive semantic graph structure, \(\odot\) denotes element-wise multiplication and \(\vee\) denotes edge-wise union.

\begin{table*}[t]
  \centering
   \caption{Evaluation results, including the AUC ($\uparrow$) scores across eight classification datasets. The best performances are highlighted in bold, and the second-best are underlined. Metric values are averaged over three random seeds. Out of In-context Window (OOW) indicates that the input length exceeds the maximum context window limit of the model and cannot be processed.}
  \resizebox{0.95\textwidth}{!}{
    \begin{tabular}{c|c|ccc|cccc|c}
    \toprule
    \multirow{2}[4]{*}{Data} & \multirow{2}[4]{*}{Shot} & \multicolumn{3}{c|}{Traditional Few-shot Tabular Learning} & \multicolumn{4}{c|}{LLM-based Few-shot Tabular Learning} & \multirow{2}[4]{*}{\textbf{\ours (Ours)}} \\
\cmidrule{3-9}          &       & SCARF & TabPFN & STUNT & In-context & TABLET & TabLLM & FeatLLM  &  \\
    \bottomrule
\multirow{3}{*}{Adult}
& 4 & $58.34_{15.42}$ & $60.89_{23.28}$ & $67.43_{29.61}$ & $77.51_{5.24}$  & $75.29_{12.24}$ & $83.57_{2.69}$ & \underline{86.68$_{0.86}$} & \textbf{86.79$_{1.78}$}\\
& 8 & $72.42_{8.95}$  & $70.42_{9.96}$  & $82.16_{6.93}$  & $79.30_{2.89}$  & $77.56_{7.56}$  & $83.52_{4.30}$ & \underline{87.89$_{0.06}$} & \textbf{87.90$_{1.65}$}\\
& 16 & $75.63_{9.56}$  & $70.34_{9.96}$  & $80.57_{10.93}$ & $79.50_{4.57}$  & $79.74_{5.64}$  & $83.23_{2.45}$ & \underline{87.54$_{0.50}$} & \textbf{88.01$_{0.98}$} \\ \midrule

\multirow{3}[0]{*}{Amazon} & 4     & 47.71\(_{4.04}\) & \underline{53.86\(_{3.38}\)} & 53.63\(_{7.99}\)  & 48.63\(_{2.68}\) & 46.71\(_{1.54}\) & 50.76\(_{8.71}\) & 48.87\(_{6.93}\) & \textbf{54.38\(_{5.17}\)} \\
& 8 & 47.71\(_{4.40}\) & \underline{54.28\(_{1.76}\)} & 54.09\(_{3.19}\)  & 48.85\(_{3.35}\) & 45.09\(_{2.03}\) & 48.53\(_{1.02}\) & 50.28\(_{6.29}\) & \textbf{54.56\(_{3.95}\)} \\
& 16 & 47.79\(_{4.01}\) & \underline{56.58\(_{4.97}\)} & 52.12\(_{0.87}\)  & 48.24\(_{4.18}\) & 49.83\(_{6.94}\) & 51.62\(_{4.16}\) & 51.33\(_{6.01}\) & \textbf{57.22\(_{5.84}\)} \\ \midrule
\multirow{3}[0]{*}{Blood} & 4 & 56.22\(_{21.00}\) & 58.72\(_{19.16}\) & 48.57\(_{6.04}\)  & 56.30\(_{12.43}\) & 56.45\(_{15.45}\) & 55.87\(_{13.49}\) & \underline{68.34\(_{7.48}\)} & \textbf{73.99\(_{1.79}\)} \\
          & 8 & 65.77\(_{5.00}\) & 66.30\(_{10.01}\) & 60.00\(_{4.84}\)  & 58.99\(_{10.12}\) & 56.37\(_{11.56}\) & 66.01\(_{9.25}\) & \underline{70.37\(_{3.23}\)} & \textbf{74.06\(_{4.45}\)} \\
          & 16 & 66.27\(_{5.04}\) & 64.14\(_{6.80}\) & 54.76\(_{4.53}\)  & 56.59\(_{5.21}\) & 60.62\(_{4.13}\) & 65.14\(_{7.55}\) & \underline{70.07\(_{5.19}\)} & \textbf{75.26\(_{0.76}\)} \\
           \midrule

\multirow{3}[0]{*}{Credit-g} & 4 & 48.92\(_{4.60}\) & 54.00\(_{7.34}\) & 48.80\(_{6.76}\)  & 52.99\(_{4.08}\) & 54.33\(_{6.54}\) & 51.90\(_{9.40}\) & \underline{55.94\(_{1.10}\)} & \textbf{60.77\(_{5.01}\)} \\
          & 8 & 55.26\(_{3.92}\) & 52.58\(_{11.27}\) & 54.50\(_{8.25}\)  & 52.43\(_{4.36}\) & 52.90\(_{5.79}\) & 56.42\(_{12.89}\) & \underline{57.42\(_{3.10}\)} & \textbf{62.94\(_{0.74}\)} \\
          & 16 & 59.22\(_{11.38}\) & 58.91\(_{8.04}\) & 57.63\(_{7.58}\)  & 55.29\(_{4.80}\) & 51.65\(_{4.02}\) & \underline{60.38\(_{14.03}\)} & 56.60\(_{2.22}\) & \textbf{69.37\(_{1.85}\)} \\
           \midrule
\multirow{3}[0]{*}{Diabetes} & 4 & 62.35\(_{7.48}\) & 56.28\(_{13.01}\) & 64.22\(_{6.78}\)  & 71.71\(_{5.31}\) & 63.96\(_{3.32}\) & 70.42\(_{3.69}\) & \underline{80.28\(_{0.75}\)} & \textbf{81.28\(_{4.05}\)} \\
          & 8 & 64.69\(_{13.33}\) & 69.08\(_{9.68}\) & 67.39\(_{12.92}\)  & 72.21\(_{2.07}\) & 65.47\(_{3.95}\) & 64.30\(_{5.88}\) & \underline{79.38\(_{1.66}\)} & \textbf{81.53\(_{1.51}\)} \\
          & 16 & 71.86\(_{3.16}\) & 73.69\(_{3.21}\) & 73.79\(_{6.48}\)  & 71.64\(_{5.05}\) & 66.71\(_{0.76}\) & 67.34\(_{2.79}\) & \textbf{80.15\(_{1.35}\)} & \underline{79.63\(_{0.36}\)} \\
          \midrule
\multirow{3}[0]{*}{Heart} & 4 & 59.38\(_{3.42}\) & 67.33\(_{15.29}\) & \underline{88.27\(_{3.32}\)}  & 60.76\(_{4.00}\) & 68.19\(_{11.17}\) & 59.74\(_{4.49}\) & 75.66\(_{4.59}\) & \textbf{89.12\(_{0.07}\)} \\
          & 8 & 74.35\(_{6.93}\) & 77.89\(_{2.34}\) & \underline{88.78\(_{2.38}\)}  & 65.46\(_{3.77}\) & 69.85\(_{10.82}\) & 70.14\(_{7.91}\) & 79.46\(_{2.16}\) & \textbf{90.65\(_{1.79}\)} \\
          & 16 & 83.66\(_{5.91}\) & 81.45\(_{5.05}\) & \underline{89.13\(_{2.10}\)} & 67.00\(_{7.83}\) & 68.39\(_{11.73}\) & 81.72\(_{3.92}\) & 83.71\(_{1.88}\) & \textbf{92.29\(_{0.74}\)} \\
          \midrule
\multirow{3}[1]{*}{Communities} & 4 & 66.18\(_{9.13}\) & OOW & 66.87\(_{14.10}\)  & OOW & OOW & OOW & \underline{75.39\(_{5.05}\)} & \textbf{75.90\(_{3.62}\)} \\
          & 8 & 72.69\(_{3.79}\) & OOW & 76.36\(_{4.55}\)  & OOW & OOW & OOW & \underline{76.59\(_{1.25}\)} & \textbf{77.12\(_{1.24}\)} \\
          & 16 & 73.09\(_{2.84}\) & OOW & 77.29\(_{2.56}\)  & OOW & OOW & OOW & 76.25\(_{0.64}\) & \textbf{78.51\(_{2.57}\)} \\
          \midrule
\multirow{3}[0]{*}{Myocardial} & 4 & 47.70\(_{4.10}\) & OOW  & 52.77\(_{2.01}\) & OOW   & OOW  & OOW   & \underline{52.87\(_{3.44}\)} & \textbf{57.46\(_{1.88}\)} \\
          & 8 & 49.37\(_{3.41}\) & OOW  & 55.40\(_{4.41}\) & OOW   & OOW   & OOW  & \underline{56.22\(_{1.64}\)} & \textbf{63.16\(_{0.87}\)} \\
          & 16 & 54.31\(_{1.42}\) & OOW   & \underline{61.22\(_{3.45}\)} & OOW   & OOW   & OOW   & 55.32\(_{9.15}\) & \textbf{63.22\(_{3.13}\)} \\
           \midrule

    \end{tabular}%
    }
  \label{tab:classification}%
\end{table*}

\subsection{Message Passing Mechanism}
\label{sec:mpm}
Through USTNE, we obtain a unified semantic node representations matrix  \(\textbf{H}\). Through Semantic Graph Structure Construction and Task-Adaptive Semantic Graph Refinement, we derive a task-adaptive semantic adjacency matrix \(\textbf{A}_{refine}\), forming a semantic graph \(\mathcal{G}=(\textbf{A}_{refine}, \textbf{H})\). To model feature interactions, we perform message passing on \(\mathcal{G}\), where the refined neighborhood of node \(v\) is defined as \(\mathcal{N}_{refine}(v)=\{u \mid \textbf{A}_{refine}(v,u)=1\}.\) Accordingly, the GNN layer is formulated as follows:
\begin{equation}
h_v^{l+1}=\sigma\Big(\textbf{W}\cdot \operatorname{AGG}({h_u^{l}:u\in \mathcal N_{refine}(v)\cup{v}})\Big),
\end{equation}
where \(h_v^{l} \in \mathbb{R}^{d}\) denotes the node representations at the \(l\)-th layer, \(\sigma\) is the ReLU activation function, \(\textbf{W} \in \mathbb{R}^{d \times d}\) is the linear transformation matrix, and AGG is neighborhood aggregation method, we use mean as aggregation method. Finally, we use a linear layer to predict the final layer output \(\textbf{H}^{\prime}=\{h_1^{\prime}, \cdots, h_n^{\prime}\}\):
\begin{equation}
\hat{y} = \text{Linear}(\text{Mean}(\textbf{H}^{\prime})),
\end{equation}
where \(\text{Mean}(\textbf{H}^{\prime})\) aggregates all node representations to produce the representation of the task, which is then passed through a linear layer for prediction. Finally, the parameters and graph structure are optimized using the following loss function:
\begin{equation}
\label{eq:loss}
    \mathcal{L} = \mathcal{L}_{task} + \lambda_1\underbrace{\mathcal{L}_{prior}}_{Eq.~\ref{eq:7}} + \lambda_2\underbrace{\mathcal{L}_{sparse}}_{Eq.~\ref{eq:15}},
\end{equation}
where, \(\lambda_1\) and \(\lambda_2\) are weighting hyperparameters that control the relative contributions of \(\mathcal{L}_{prior}\) and \(\mathcal{L}_{sparse}\) (w.r.t. \(\mathcal{L}_{task}\)), thereby balancing task fitting, prior learning, and sparsity regularization, $\mathcal{L}_{task}$ denotes the task loss, which depends on the downstream setting. For classification, we use the cross-entropy loss
$\mathcal{L}_{task} = -\sum_{c} y_c \log \hat{y}_c$,
and for regression, we use the squared error
$\mathcal{L}_{task} = \|y-\hat{y}\|_2^2$. \(\mathcal{L}_{sparse}\) is a sparsity regularizer to encourage a sparse refined structure as follows:
\begin{equation}
\label{eq:15}
\mathcal{L}_{sparse} = \sum_{(v,u):\,\textbf{A}_{refine}(v,u)=1} s_{v,u},
\end{equation}
where \(s_{v,u}\) is the semantic score of the edge \((v, u)\). Refer to Appendix~\ref{suba:alg} for the detailed algorithm.

\section{Experiments}

\textbf{Datasets.} We evaluate \ours\ on eleven real-world datasets, including eight for classification tasks: Adult~\cite{asuncion2007uci}, Amazon~\cite{rani2023amazon}, Blood~\cite{yeh2009knowledge}, Credit-g~\cite{kadra2021well}, Diabetes\footnote{\url{https://www.kaggle.com/datasets/uciml/pima-indians-diabetes-database}}, Heart\footnote{\url{https://www.kaggle.com/datasets/fedesoriano/heart-failure-prediction}}, Communities~\cite{asuncion2007uci}, and Myocardial~\cite{myocardial_infarction_complications_579}, and three for regression tasks from OpenML~\cite{vanschoren2014openml}, i.e., Abalone, Boston, and Cholesterol. Dataset statistics details are provided in Appendix~\ref{suba:datasets}.\\
\textbf{Baselines.} 
We compare \ours\ against a broad set of baselines for few-shot tabular learning. These include Traditional few-shot tabular learning methods: SCARF~\cite{bahri2021scarf}, TabPFN~\cite{hollmann2022tabpfn}, and STUNT~\cite{nam2023stunt}, and LLM-based few-shot methods: In-context~\cite{wei2022emergent}, TABLET~\cite{slack2023tablet}, TabLLM~\cite{hegselmann2023tabllm}, and FeatLLM~\cite{featllm}. In addition, we evaluate different graph construction strategies: TabGSL~\cite{liao2023tabgsl} and T2G-FORMER~\cite{yan2023t2g}, to assess the effectiveness of the graph structure constructed by \ours.\\
\textbf{Implementation Details}. We use the BERT~\cite{devlin2019bert} model’s embedding layer as the text encoder and GPT-4o-mini~\cite{achiam2023gpt} for generating graph structures. Topk is set to 5, $\tau$ is set to 0.2, $\lambda_1$ and $\lambda_2$ are set to 0.1 for task-adaptive semantic graph refinement. In the API, we set the temperature for inference by the LLM to 0.0 and kept the top-\(p\) value at its default setting of 1. For training, we utilize the Adam optimizer with a learning rate of 1e-5, and the model is trained for 1000 iterations. For evaluation, we use the Area Under the Curve (AUC) metric for classification tasks and Root Mean Squared Error (RMSE) for regression tasks.

\subsection{Main Results}
\begin{figure}[t]
\centering
\includegraphics[width=0.48\textwidth]{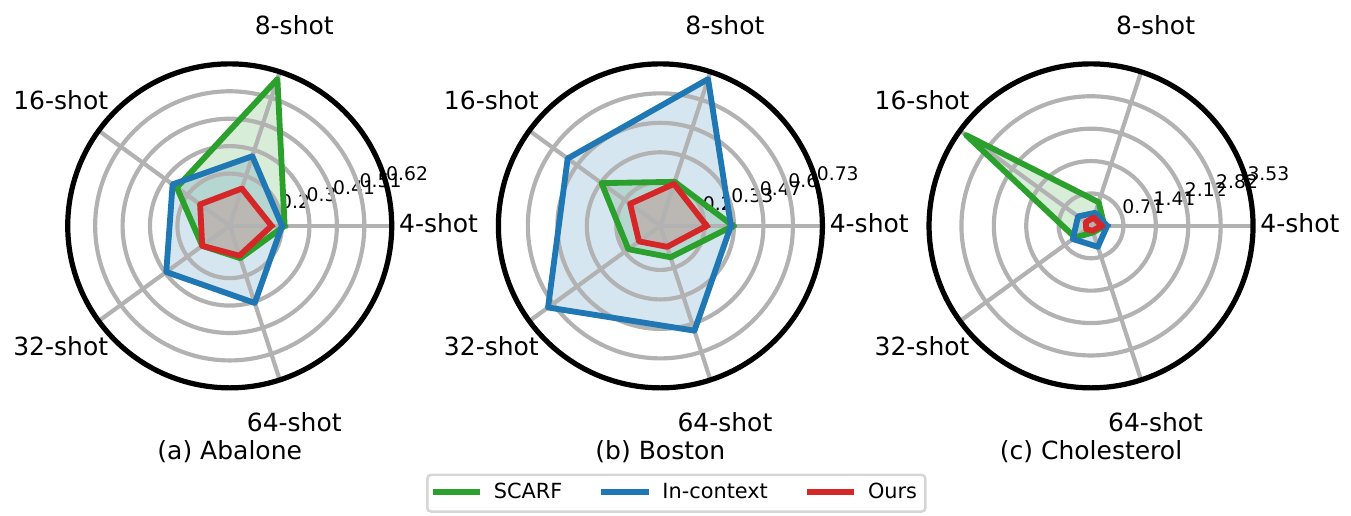}
\caption{\label{figure:radar} Evaluation results, including the RMSE ($\downarrow$) scores across three regression datasets. TabPFN, STUNT, TABLET, TabLLM, and FeatLLM are specifically designed for few-shot classification tasks and are not suitable for regression tasks.
}
\end{figure}

\begin{table}[t]
  \centering
  \caption{Runtime in seconds of \ours\ and other few-shot tabular baseline methods for the training and inference phase, conducted on the Myocardial dataset.}
    \begin{tabular}{c|ccc}
    \toprule
    \multicolumn{2}{c|}{Model} & \multicolumn{1}{l}{Training} & \multicolumn{1}{l}{Inference} \\
    \midrule
    \multirow{3}[2]{*}{Traditional } & \multicolumn{1}{l|}{SCARF} & 853.718       & 0.025 \\
          & \multicolumn{1}{l|}{TabPFN} &  0.880    & 4.902 \\
          & \multicolumn{1}{l|}{STUNT} &  1285.592     & 0.031 \\
    \midrule
    \multirow{3}[2]{*}{LLM-based }
          & \multicolumn{1}{l|}{TABLET} &  1.626     & 2232.55 \\
          & \multicolumn{1}{l|}{TabLLM } &  502.484     &  1429.87\\
          & \multicolumn{1}{l|}{FeatLLM} &  1720.188     & 0.032  \\
    \midrule
    \multicolumn{2}{c|}{\textbf{\ours\ (Ours)}} & \textbf{0.862}      & \textbf{0.023} \\
    \bottomrule
    \end{tabular}%
  \label{tab:eff}%
\end{table}%
\textbf{Q: Does our method perform well in few-shot tabular prediction?} Yes, as shown in Tab.~\ref{tab:classification} and Fig.~\ref{figure:radar}, our method achieves SOTA performance on both classification and regression tasks.\\
$\rhd$ \textsf{Classification Task Comparison.} The comparison results are summarized in Tab.~\ref{tab:classification}. From this, we can conclude that our method outperforms other few-shot methods in few-shot tabular classification tasks. As shown in Tab.~\ref{tab:classification}, our method achieves better results than the best few-shot tabular classification baseline method FeatLLM and reaches state-of-the-art performance on this task. This result demonstrates that introducing a semantic graph structure inferred from LLM and combining it with task-adaptive structural refinement can more effectively capture task-related semantic dependencies between features, thereby improving classification performance in few-shot scenarios. \ding{182} \textit{The LLM-based method significantly outperforms Traditional models in few-shot scenarios.} The results in Tab.~\ref{tab:classification} show that the LLM-based method significantly outperforms Traditional models. This is mainly due to the fact that LLM can utilize its real-world knowledge learned during the pre-training stage to perform stronger semantic understanding and reasoning on tabular data, while traditional methods usually rely only on limited labeled signals in few-shot scenarios, making it difficult to obtain sufficient semantic inductive ability. Unlike directly using LLM to reason about the entire table, \ours\ only uses feature names to prompt LLM infer the semantic graph, then refines and aligns it with the task through a task adaptation mechanism, thereby injecting effective structural prior information in a lighter way. \\
$\rhd$ \textsf{Comparison in Regression Tasks}. We further evaluate \ours\ on three few-shot regression datasets. As shown in Fig.~\ref{figure:radar}, \ding{183} \textit{\ours\ not only stably applies to regression tasks but also achieves state-of-the-art performance.} This demonstrates that the semantic graph structure constructed by \ours\ can also provide effective structural priors for regression. Compared to the existing few-shot tabular learning baselines, our method achieves state-of-the-art (SOTA) performance on few-shot regression tasks. Overall, these results validate that graph priors inferred by LLM can still capture semantic dependencies between features in regression settings, thereby improving prediction accuracy.\\

\noindent \textbf{Q: Does \ours\ have a high computational efficiency?} Yes, as shown in Tab.~\ref{tab:eff}, \ours\ have a high computational efficiency in both training and inference. This efficiency arises from two key design choices: (i) Compared to Traditional few-shot tabular learning methods that rely on additional pre-training with unlabeled or synthetic data, \ours\ requires no additional pre-training phase, thus significantly reducing overall training overhead; (ii) Compared to LLM-based few-shot tabular methods, \ours\ avoids performing LLM queries for each instance to complete predictions. Instead, we obtain the graph structure inferred by LLM as a structural prior with just a single API call, which can be reused in subsequent training and inference, thereby utilizing LLM resources more efficiently and reducing inference costs.
\begin{figure}[t]
\centering
\includegraphics[width=0.47\textwidth]{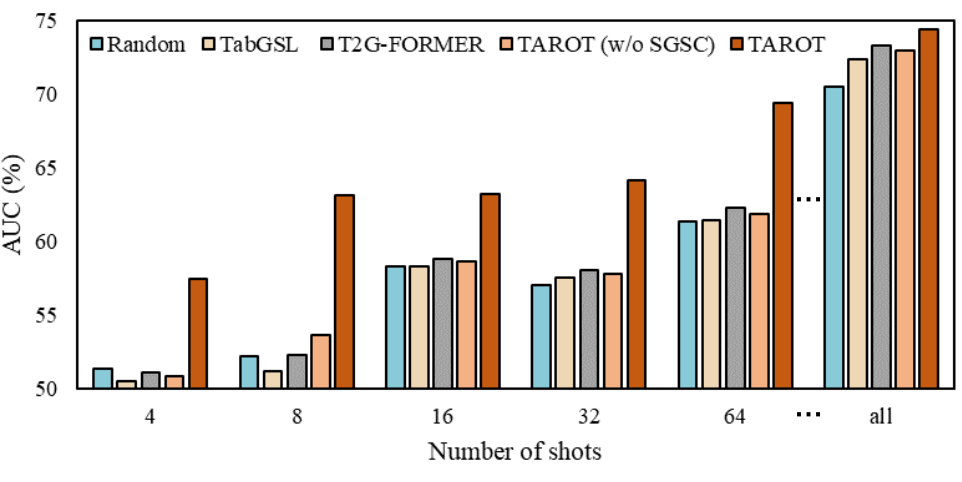}
\caption{\label{figure:fulldata} Quantitative analysis of the effectiveness of generating graph structures on Myocardial. w/o SGSC: remove \underline{S}emantic \underline{G}raph \underline{S}tructure \underline{C}onstruct and refine an empty prior graph via Task-Adaptive Semantic Graph Refinement. To fairly compare the five methods differ only in their graph structures.
}
\end{figure}

\subsection{Effectiveness of Graph Structure}
\begin{figure}[htbp]
\centering
\includegraphics[width=0.48\textwidth]{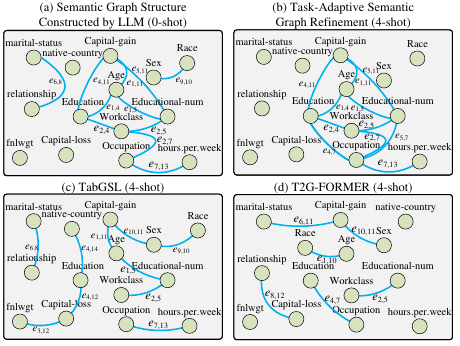}
\caption{\label{figure:graph} Qualitative analysis of the effectiveness of generating graph structures on Adult, the task is ``Predict whether the person earns more than 50000 dollars per year?''
}
\end{figure}
\noindent \textbf{Q: Dose \ours\ generated an effective graph structure?} Yes, we analyze the effectiveness of the graph structure generated by \ours\ from both quantitative and qualitative perspectives, as shown in Fig.~\ref{figure:fulldata} and~\ref{figure:graph}, respectively.\\
$\rhd$ \textsf{Quantitative analysis}. The quantitative analysis of the effectiveness of various graph construction strategies shown in Fig.~\ref{figure:fulldata} in downstream tasks, we have two observations: \ding{184} \textit{Previous graph construction strategies fail to capture semantic relationships between features in few-shot scenarios.} As shown in Fig.~\ref{figure:fulldata}, in the few-shot scenarios (e.g., 4-64 shots), the graph structures learned by the previous methods (TabGST and T2G-FORMER) perform similarly to the randomly generated graphs in downstream tasks, indicating that these methods have difficulty learning effective graph structures in few-shot settings. In contrast, \ours\ utilizes the knowledge from LLM to construct a semantic graph structure and further performs task-adaptive semantic graph refinement, resulting in an effective graph structure that promotes the performance in downstream tasks. \ding{185} \textit{When Semantic Graph Structure Construction (SGSC) is removed (i.e., w/o SGSC), our method degenerates to conventional graph structure learning methods.} As shown in Fig.~\ref{figure:fulldata}, removing SGSC causes performance in few-shot settings to drop to the level of randomly generated graph structures, suggesting that the semantic graph produced by SGSC provides useful priors for constructing effective graph structures.

\noindent$\rhd$ \textsf{Qualitative analysis}. The qualitative analysis of the effectiveness of various graph construction strategies for downstream tasks, shown in Fig.~\ref{figure:graph}, \ding{186}\textit{in few-shot scenarios, \ours\ can effectively induce task-related semantic graph structures.} Specifically, as illustrated in Fig.~\ref{figure:graph}, \ours\ first elicits a meaningful structure prior via a zero-shot LLM prompt, and then suppresses spurious connections through Task-Adaptive Semantic Graph Refinement. For example, it removes the task-unrelated edge between sex and race when predicting whether income exceeds $\$50{,}000$. In contrast, the graphs produced by TabGSL and T2G-FORMER under the 4-shot regime are close to random, containing substantial structural noise, which in turn hampers their ability to capture reliable feature dependencies and degrades downstream performance.

\subsection{Further Analysis}
\label{sub:as}
\begin{table}[t]
  \centering
  \caption{Effect of model components on Myocardial. USTNE denotes \underline{U}nified \underline{S}emantic \underline{T}abular \underline{N}ode \underline{E}ncoder, CT denotes \underline{C}ode \underline{T}emplate, TAP denotes \underline{T}ask-\underline{A}daptive \underline{P}runing, TAE denotes \underline{T}ask-\underline{A}daptive \underline{E}nhancement}
  \resizebox{0.47\textwidth}{!}{
    \begin{tabular}{c|ccccc|c}
    \toprule
    \multirow{2}[4]{*}{Ablation} & \multicolumn{5}{c|}{Shot}              & \multirow{2}[4]{*}{Avg} \\
\cmidrule{2-6}          & 4     & 8     & 16    & 32    & 64    &  \\
    \midrule
    \multicolumn{7}{c}{Node Embedding} \\
    w/o USTNE & 54.58 & 61.27 & 61.36 & 62.03 & 65.89 & 61.03 \\
    \midrule
    \multicolumn{7}{c}{Semantic Graph Structure Construction} \\
    w/o Instruction & 51.89 & 54.11 & 59.68 & 63.28 & 62.72 & 58.34\\
    w/o CT & 51.52 & 53.85 & 59.02 & 62.59 & 62.29 & 57.85 \\

    \midrule
    \multicolumn{7}{c}{Task-Adaptive Semantic Graph Refinement} \\
    w/o TAE & 56.12 & 61.89 & 62.05 & 62.91 & 65.99 & 61.79 \\
    w/o TAP & 55.78 & 61.53 & 61.77 & 62.64 & 65.57 & 61.46 \\
    w/o $\mathcal{L}_{sqarse}$ & 55.87 & 61.57 & 61.89 & 62.34 & 65.37 & 61.40 \\
    w/o $\mathcal{L}_{prior}$  & 50.78& 53.78 & 58.57 & 57.93 & 62.15 & 56.64 \\
    \midrule
    \textbf{\ours\ (Ours)}  & \textbf{57.46} & \textbf{63.16} & \textbf{63.22} & \textbf{64.12} & \textbf{69.37} & \textbf{63.47} \\
    \bottomrule
    \end{tabular}%
    }
  \label{tab:ablation}%
\end{table}%

\begin{figure}[t]
\centering
\includegraphics[width=0.49\textwidth]{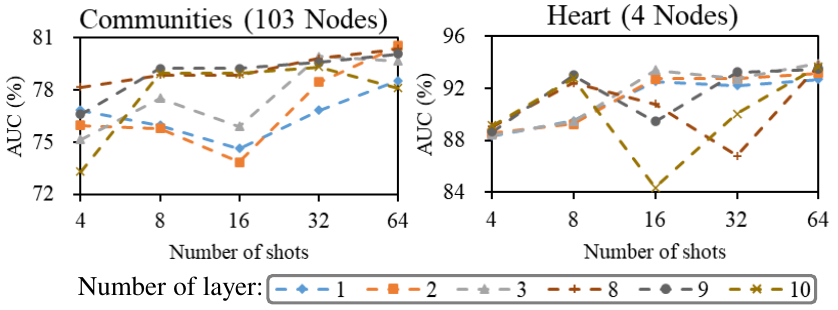}
\caption{\label{figure:per} Impact of varying the number of GNN layers on the performance of \ours\ across different datasets.
}
\end{figure}

\textbf{Q: Does each component of \ours\ contribute effectively to the overall performance?} Yes, each component of \ours\ contribute effectively to the overall performance.\\
$\rhd$ \textsf{Ablation Analysis.} As shown in Tab.~\ref{tab:ablation}, \ding{187}  \textit{the prior learning loss $\mathcal{L}_{prior}$ constrains Task-Adaptive Semantic Graph Refinement.} Removing $\mathcal{L}_{prior}$ leads to a significant performance drop, indicating that the prior learning loss $\mathcal{L}_{prior}$ constrains the Task-Adaptive Semantic Graph Refinement by referencing the structural prior in the semantic graph generated by the LLM, rather than arbitrarily adding or pruning semantic edges. \\ 
\noindent$\rhd$ \textsf{Analysis of the Relationship between GNN Layers and Node Size.} As shown in Fig.~\ref{figure:per}, \ding{188}  \textit{datasets with fewer nodes typically benefit from shallower GNNs, while larger graphs require deeper GNNs.} For small-node datasets (e.g., Heart with 11 nodes), adding more GCN layers makes performance more sensitive to the number of nodes and can be hurt due to over-smoothing and information loss. In contrast, for large-node datasets (e.g., Myocardial with 111 nodes), deeper GNNs better capture complex feature interactions and global structure, leading to improved performance.
\section{Conclusion}
In this paper, we introduced \ours, a task-adaptive GNN-based framework for few-shot tabular learning that explicitly models feature interactions through semantic graphs. 
Motivated by the observation that existing few-shot tabular learning paradigms largely overlook semantic relationships between features, \ours\ leverages LLMs to induce a semantic graph based on task descriptions and feature names, providing a strong structural prior under limited supervision. To address the inevitable structural noise introduced by LLM inference, we further proposed a task-adaptive graph refinement mechanism that prunes spurious and task-unrelated edges while recovering missing task-related semantic ones. 
Extensive experiments on 11 real-world datasets demonstrate that \ours\ consistently outperforms state-of-the-art Traditional and LLM-based baselines in few-shot settings. 
\section{ACKNOWLEDGMENTS}
This work was supported by a grant from the National Natural Science Foundation of China under grants (No.62372211), and the Science and Technology Development Program of Jilin Province (No.20250102216JC).

\bibliographystyle{ACM-Reference-Format}
\bibliography{sample-base}

\appendix
\section{Algorithm of \ours}
\label{suba:alg}

\tcbset{
    colback=white!10,       
    colframe=black,        
    boxrule=0.3mm,         
    arc=0.5mm,               
    fonttitle=\bfseries,   
    left=2mm, right=2mm,   
    top=1mm, bottom=1mm    
}
\begin{table}[h]
 \centering
  \caption{\label{figure:prompt}Prompt \(p\) used by \ours\ of Adult dataset.}
\begin{tcolorbox}
     Meta Information $p_{meta}$\\
    \textcolor[rgb]{ .929,  .49,  .192}{\textcolor[rgb]{ .0,  .0,  .0} {Task objective}: Does this person earn more than 50000 dollars per year?\\
    \textcolor[rgb]{ .0,  .0,  .0}{features descriptions}: age: the age of an individual, workclass: a general term to represent the employment status of an individual, fnlwgt: the number of units in the target population that the responding unit represents, education: the highest level of education achieved by an individual, educational-num: the highest level of education achieved in numerical form... } \\
    \\
    Instruction  $p_I$  \\
    \textcolor[rgb]{ .494,  .392,  .62}{Step 1. Analyze all the causal relationships or tendencies between features based on general knowledge and common sense within a short sentence to answer the task.\\
    Step 1. Based on the above tabular description and Step 1’s results, analyze the semantic relationships between features}\\ \\
Code Template $p_{code}$\\
\textcolor[rgb]{ 1.0,  .4,  .4}
{
import numpy as np\\
feature\_number = \(n\)\\
A = np.zeros((feature\_number, feature\_number), dtype=int)\\
\# \textit{semantic relationships between features}\\
edges = [\textbf{Filled by LLM}]\\
for i, j in relationships:\\
    \text{\ \ \ A[i, j] = A[j, i] = 1}\\
np.save(graph\_path, A)
}
\end{tcolorbox}

\label{tab:adult}%
\end{table}%

\begin{algorithm}
\caption{\ours}
\begin{algorithmic}[1]
\State Given a tabular dataset $D=\{\mathbb{X},\mathbb{Y}\}$
\State For \(k\)-shot learning, we construct the training set $\hat{D}$ by sampling
\(k\) labeled samples from the dataset $D$.
\State $h_{i}=\text{USTNE}(x_{i}, f_j),\ \ h_{i} \in \textbf{H},x_i \in \hat{D}$
\State Use the formula: $\textbf{A}= \text{Execute}(\text{LLM}(p))$ to obtain the semantic graph structure $\textbf{A}$, the prompt \(p\) used are shown in Tab.~\ref{figure:prompt}.
\State \textbf{repeat}
\State \quad $e_{v,u}=\big[(h_v,h_u) \oplus (h_v\odot h_u)\oplus(|h_v-h_u|)\big]$
\State \quad $s_{v,u} = \sigma(e_{v,u}w)$
\State \quad Obtain $\textbf{A}_{prune}$ and $\textbf{A}_{enhance}$ according to Eq.~\ref{eq:prune} and Eq.~\ref{eq:enhance}, respectively
\State \quad $\textbf{A}_{refine} = (\textbf{A}_{prior} \odot (1-\textbf{A}_{prune})) \lor \textbf{A}_{enhance}$

\State \quad \textbf{for} $l = 1, \dots, L$ \textbf{do}
\State \quad \quad $h_v^{l+1}=\sigma\Big(W\cdot \operatorname{AGG}({h_u^{l}:u\in \mathcal N_{refine}(v)\cup{v}})\Big),$
\State \quad \textbf{end for}
 \State \quad $\hat{y} = \text{Linear}(\text{Mean}(H^L))$
\State \quad Calculate the loss according to Eq.~\ref{eq:loss}
\State \quad Update parameters using Adam optimizer
\State \textbf{until} converges
\end{algorithmic}
\label{alg:train}
\end{algorithm}
Given a labeled tabular dataset \(D={\mathbb{X},\mathbb{Y}}\), we first construct a \(k\)-shot training set \(\hat{D}\) by sampling \(k\) labeled instances per task (or per class). For each training instance, \ours\ applies the \emph{Unified Semantic Tabular Node Encoder} (USTNE) to transform heterogeneous features into unified semantic node representations, producing a node embedding matrix \textbf{H}. Next, to obtain a semantics-aware prior structure without accessing any raw samples, we query LLMs with a prompt \(p\) that contains task meta-information, step-by-step reasoning instructions, and an executable code template; executing the returned code directly yields the semantic adjacency matrix \(\mathbf{A}\), which is treated as the prior graph \(\mathbf{A}_{prior}\).

During training, \ours\ performs \emph{task-adaptive semantic graph refinement} to mitigate structural noise introduced by the LLM. Specifically, for each node pair \((v,u)\), we construct an edge representation \(e_{v,u}\) by concatenating \((h_v,h_u)\), element-wise product \((h_v\odot h_u)\), and absolute difference \(|h_v-h_u|\), and obtain a semantic score \(s_{v,u}=\sigma(e_{v,u}w)\) via a linear scorer. Based on these scores, we derive a pruning mask \(\mathbf{A}_{prune}\) (removing edges with scores below a threshold \(\tau\) and an enhancement mask \(\mathbf{A}_{enhance}\) (adding top-(k) highest-scoring edges), and refine the structure by
\(\mathbf{A}_{refine}=(\mathbf{A}_{prior}\odot(1-\mathbf{A}_{prune}))\lor \mathbf{A}_{enhance}\).
We then perform message passing on the refined graph for \(L\) GNN layers using mean aggregation over the refined neighborhood \(\mathcal{N}_{refine}(v)\), producing \(\mathbf{H}^L\). The final prediction is obtained by mean-pooling node embeddings followed by a linear head, i.e., \(\hat{y}=\mathrm{Linear}(\mathrm{Mean}(\mathbf{H}^L))\). All parameters (including the edge scorer and GNN) are optimized end-to-end with Adam by minimizing
\(\mathcal{L}=\mathcal{L}_{task}+\lambda_1\mathcal{L}_{prior}+\lambda_2\mathcal{L}_{sparse}\),
where \(\mathcal{L}_{prior}\) regularizes the learned structure toward \(\mathbf{A}_{prior}\) and \(\mathcal{L}_{sparse}\) encourages sparsity in \(\mathbf{A}_{refine}\), improving stability and generalization in few-shot tabular learning.
\section{Detailed Experiment Setups}
\subsection{Datasets}
\label{suba:datasets}
\begin{table}[htbp]
  \centering
  \caption{The basic information of each dataset used in our experiments.}
  \resizebox{0.47\textwidth}{!}{
    \begin{tabular}{cccccc}
    \toprule
     Dataset & \# of samples & \# of features & \# of cat & \# of num & Task type \\
    \midrule
     Adult & 48842 & 14    & 7     & 7 &classification\\
    Amazon & 2000  & 9     & 9     & 0 &Classification\\
          Blood & 748   & 4     & 0     & 4 &Classification\\
        Credit-g & 1000  & 20    & 12    & 8 &Classification\\
          Diabetes & 768   & 8     & 0     & 8 &Classification\\
          Heart & 918   & 11    & 4     & 7 &Classification\\
           Communities & 1994  & 103   & 1     & 102 &Classification\\
          Myocardial & 1700  & 111   & 94    & 17 &Classification\\
          Abalone & 4177  & 8   & 1    & 7 &Regression\\
          Boston & 506  & 13   & 2    & 11 &Regression\\
          Cholesterol & 303  & 9   & 4    & 17 &Regression\\
    \bottomrule
    \end{tabular}%
    }
  \label{tab:dataset}%
\end{table}%
We evaluate \ours\ on 11 tabular benchmarks spanning both classification and regression, with diverse feature heterogeneity (mixed categorical and numerical attributes). Specifically, we use 8 classification datasets: Adult (48,842 samples, 14 features), Amazon (2,000, 9), Blood (748, 4), Credit-g (1,000, 20), Diabetes (768, 8), Heart (918, 11), Communities (1,994, 103), and Myocardial (1,700, 111); and 3 regression datasets: Abalone (4,177, 8), Boston (506, 13), and Cholesterol (303, 9). Overall, dataset sizes range from 303 to 48,842 and feature dimensionalities from 4 to 111, with substantial variation in the categorical/numerical composition. For example, Amazon contains only categorical features (9/0), Blood and Diabetes contain only numerical features (0/4 and 0/8), Communities is high-dimensional and predominantly numerical (1/102), whereas Myocardial is categorical-dominant (94/17). This diversity in task type, scale, and feature modality enables a comprehensive assessment of few-shot tabular learning performance. 

\subsection{Baselines} 
\label{suba:baselines}
We compare \ours\ with a broad set of representative baselines for few-shot tabular learning, covering (i) traditional few-shot tabular learners, (ii) LLM-based few-shot approaches, and (iii) graph-based tabular learning methods for semantic structure construction. For a fair comparison, all methods are evaluated under the same $k$-shot protocol (i.e., sampling $k$ labeled instances to form the support/training set). Whenever official implementations are available, we use them and keep the default hyperparameters; otherwise, we follow the settings recommended in the original papers.

\noindent\textbf{$\triangleright$ Traditional few-shot tabular learning methods.}
These methods improve few-shot generalization by leveraging transferable inductive biases learned from additional pretraining or self-supervision on tabular data.
\begin{itemize}
    \item \textbf{SCARF}~\cite{bahri2021scarf} is a self-supervised contrastive pretraining method that generates two views by randomly corrupting features and learns invariant representations by aligning original and corrupted samples.
    \item \textbf{TabPFN}~\cite{hollmann2022tabpfn} pretrains a transformer on a large collection of synthetic tabular tasks, enabling strong few-shot predictions via a single forward pass without dataset-specific hyperparameter tuning.
    \item \textbf{STUNT}~\cite{nam2023stunt} is a few-shot semi-supervised framework that constructs pseudo-tasks and leverages self-training style transfer to improve performance with limited labels.
\end{itemize}

\noindent\textbf{$\triangleright$ LLM-based few-shot tabular learning methods.}
These methods serialize tabular samples into natural language (or structured text) and prompt LLMs to perform prediction, leveraging the world knowledge and reasoning abilities encoded in LLMs.
\begin{itemize}
    \item \textbf{In-context}~\cite{wei2022emergent} is a standard prompting baseline that provides a few labeled examples together with the test instance, allowing the LLM to infer the mapping via in-context learning without parameter updates.
    \item \textbf{TABLET}~\cite{slack2023tablet} improves LLM-based tabular reasoning by incorporating task-specific instructions and structured prompts on top of vanilla in-context learning.
    \item \textbf{TabLLM}~\cite{hegselmann2023tabllm} adapts LLMs to tabular prediction by fine-tuning on tabular-formatted data, enhancing their awareness of feature-value semantics.
    \item \textbf{FeatLLM}~\cite{featllm} uses LLMs as feature engineers to identify and filter informative features (or generate feature rationales), and then trains downstream predictors on the resulting feature subset for few-shot learning.
\end{itemize}

\noindent\textbf{$\triangleright$ Graph-based tabular learning / graph construction strategies.}
To assess the benefit of explicitly modeling semantic feature dependencies, we additionally compare methods that construct feature graphs and perform relational reasoning.
\begin{itemize}
    \item \textbf{TabGSL}~\cite{liao2023tabgsl} learns a graph structure by initializing candidate connections (e.g., via similarity/kNN) and retaining top-$k$ relations to form a learner-view graph for message passing.
    \item \textbf{T2G-FORMER}~\cite{yan2023t2g} computes feature relationship scores to induce a graph structure and models feature dependencies using transformer-style graph reasoning.
\end{itemize}
\section{Additional Experiments}
\begin{table}[t]
\centering
\caption{Missing column name experiments on Diabetes, averaged over 4, 8, and 16 shots. The missing rate denotes the percentage of feature names/descriptions removed from the metadata.}
\label{tab:missing_metadata}
\small
\setlength{\tabcolsep}{5pt}
\begin{tabular}{lcccc}
\toprule
\textbf{Method} & \multicolumn{4}{c}{\textbf{Missing rate (\%)}} \\
\cmidrule(lr){2-5}
 & \textbf{0} & \textbf{10} & \textbf{50} & \textbf{100} \\
\midrule
STUNT   & 68.47 & 68.47 & 68.47 & 68.47 \\
FeatLLM & 79.94 & 77.44 & 71.32 & 66.89 \\
\ours   & 80.81 & 78.85 & 72.24 & 67.74 \\
\ours\_S & \textbf{80.85} & \textbf{79.58} & \textbf{73.89} & \textbf{70.21} \\
\bottomrule
\end{tabular}
\end{table}
\noindent\textbf{Analysis of Missing Metadata Robustness.}
As shown in Table~\ref{tab:missing_metadata}, \ours\ consistently outperforms FeatLLM under different missing-metadata rates, that semantic graph provide useful prior knowledge in few-shot settings. Although the performance of metadata-based methods decreases as more feature names/descriptions are removed, \ours\ remains more robust than FeatLLM across all settings. When all metadata is missing, \ours\_S, which uses statistical dependency signals such as Pearson's $R$, correlation ratio, and Cramér's $V$, achieves 70.21 AUC and improves \ours\ by 2.47 points. These results show that dependency-based auxiliary signals can effectively mitigate the degradation caused by missing semantic information.

\end{document}